\documentclass[runningheads]{llncs}
\usepackage[T1]{fontenc}
\usepackage{newtxtext}
\usepackage[varvw]{newtxmath}
\usepackage{graphicx}
\usepackage{booktabs}
\usepackage{multirow}
\usepackage{tabularx}
\usepackage{makecell}
\usepackage{wrapfig}
\usepackage{float}
\usepackage[table]{xcolor}
\usepackage{siunitx}
\usepackage{flafter}
\usepackage{marvosym}

\definecolor{propfill}{HTML}{E6F1FB}
\definecolor{redgood}{HTML}{0F6E56}

\begin{document}

\title{Towards Trustworthy AI-Enabled Emergency Medical Dispatch:
Uncertainty-Aware Ambulance Travel Time Estimation}

\titlerunning{Uncertainty-Aware Ambulance Travel Time Estimation}

\author{Muhammad Adil Abid\,\textsuperscript{\Letter}\inst{1,2}
\orcidID{0000-0002-0403-5353}
\thanks{Corresponding author: \email{muhammad.adil-abid@mau.se}} \and
Johan Holmgren\inst{1,2}\orcidID{0000-0001-7773-9944} \and
Fabian Lorig\inst{1,2}\orcidID{0000-0002-8209-0921} \and
Jesper Petersson\inst{3,4}\orcidID{0000-0003-3322-6383}}

\authorrunning{M. A. Abid et al.}

\institute{
Department of Computer Science and Media Technology,
Malmö University, 21119 Malmö, Sweden \and
Sustainable Digitalisation Research Centre, Malmö University, Sweden \and
Department of Health Care Management, Region Skåne, 21428 Malmö, Sweden \and
Department of Neurology, Lund University, 22242 Malmö, Sweden \\
\email{\{muhammad.adil-abid,johan.holmgren,fabian.lorig\}@mau.se}\\
\email{jesper.petersson@skane.se}
}

\maketitle

\begin{abstract}
Emergency medical dispatch increasingly relies on AI-enabled decision support
systems when making ambulance dispatch decisions. In this process, accurate
ambulance travel time estimates are essential for timely resource allocation
and prehospital care. However, existing ambulance travel time estimation
approaches typically produce only single point predictions. Consequently, they
do not indicate how reliable an individual estimate is or how much uncertainty
it carries. This limitation is particularly problematic in safety-critical
emergency medical service dispatch operations, where trips with similar
expected travel times may exhibit substantially different levels of
uncertainty. Furthermore, real-world ambulance trip logs may contain
domain-specific artefacts, such as co-located healthcare facilities and
physically implausible travel speeds. These artefacts can degrade point
prediction accuracy and affect the prediction task on which subsequent
uncertainty estimation depends. To address these challenges, we propose an
uncertainty-aware methodology for ambulance travel time estimation and
evaluate it through a retrospective real-world scenario study using two years
of ambulance trip data from the Skåne region in Sweden. The proposed
methodology moves beyond average-case point predictions by generating
context-adaptive prediction intervals with stable empirical coverage across
the evaluated operationally important contexts. The resulting prediction
intervals provide explicit information about predictive uncertainty that can
support more transparent and risk-sensitive ambulance dispatch decision-making.

\end{abstract}

\section{Introduction}
\label{sec:introduction}
\vspace{-5pt}
Emergency medical dispatch is a high-stakes decision-making process that
increasingly relies on AI-enabled decision support systems that combine
operational data, predictive models, and human judgement. Within this
public-sector service, ambulance travel time estimation is an important
component of dispatch decision support, as travel time estimates provide
important information on which available unit should be assigned to an
incident and how quickly prehospital care may reach the patient.
Even small estimation errors can lead to the assignment of an ambulance that
arrives later than another available unit, which matters most for
high-priority incidents where delays may have clinical consequences.
Ambulance travel time estimation is challenging because travel times vary with
local conditions, dispatch priority, route characteristics, and temporal
context~\cite{introduction-ambulance-travel-time-patterns}. To capture this
complexity, recent research has increasingly adopted data-driven machine
learning approaches that can improve predictive accuracy compared with static
lookup tables, fixed-speed assumptions, and generic navigation
estimates~\cite{related-work-Google-and-Data-driven-ambulance-travel-time-estimation,introduction-Adil-ambulance-travel-time,introduction-ambulance-travel-time-google-maps-estimation}.

However, predictive accuracy alone is insufficient for safety-critical
dispatch support. Most existing ambulance travel time estimation approaches
provide only a single point prediction. A single point prediction provides one
estimated travel time, for example, that an ambulance will arrive in
$12$~minutes, without indicating the uncertainty associated with that
estimate. This missing information is important in emergency medical service
(EMS) decision support because predictions with similar expected travel times
may differ substantially in their uncertainty. For example, consider two
ambulances available for a time-critical emergency. A point prediction model
estimates travel times of $10$~minutes for Ambulance~A and $11$~minutes for
Ambulance~B, which would favour Ambulance~A based on the expected travel time
alone. Suppose, however, that an uncertainty-aware methodology produces an interval
of $\pm 5$~minutes for Ambulance~A and $\pm 1$~minute for Ambulance~B. If an
operational criterion requires arrival within $12$~minutes, Ambulance~B may
represent the lower-uncertainty option under that criterion. More generally,
uncertainty information could support dispatchers when considering whether an
individual travel time estimate is sufficiently reliable or whether
additional operational information should be considered. Thus, an
uncertainty-aware estimation complements the point estimate by explicitly
representing the range of plausible travel times rather than providing only an
expected value. As our exploratory analysis in Section~\ref{sec:EDA} shows,
travel time variability differs across temporal conditions and route frequency
groups, while substantial variability remains even among trips with similar
operational characteristics.

Another important challenge concerns the operational validity of the
\emph{underlying data} on which predictive models should be developed.
Real-world EMS datasets may contain domain-specific artefacts that distort the
prediction task. First, operationally distinct facilities, such as co-located
hospitals and on-campus clinics, may be geocoded to identical coordinates,
creating ambiguity in origin--destination representations. Second, a small
subset of trips may have implied average speeds outside physically plausible
ranges for ground ambulance transport. Extremely low speeds may indicate idle
or waiting time included in the recorded trip duration, whereas implausibly
low or high speeds may also result from GPS inaccuracies or timestamp
inconsistencies. For example, departure or arrival events may be manually
registered before or after the corresponding operational event, making the
recorded trip duration too long or too short. We therefore define two curation
criteria a priori from domain knowledge: removing trips between facilities
that share the same coordinates and removing trips with implausible average
speeds. Applying these criteria before model development provides a stronger
basis for point prediction and subsequent uncertainty estimation.

In this paper, we focus on three limitations that remain insufficiently
addressed in ambulance travel time estimation research:
(i)~\emph{deterministic predictions}, which do not quantify the uncertainty
associated with individual travel time estimates;
(ii)~\emph{operational data quality}, where domain-specific artefacts in
historical dispatch records may distort the prediction task; and
(iii)~\emph{subgroup reliability}, where satisfactory aggregate calibration
may conceal differences in empirical coverage across operationally relevant
contexts. To address these limitations, we propose an uncertainty-aware methodology
for ambulance travel time estimation and evaluate it through a retrospective
real-world scenario study using two years of ambulance trip data from the
Skåne region in Sweden. The proposed methodology first applies domain-informed data
curation to establish an operationally valid prediction task. The resulting
data are then prepared for model development, and a point prediction model is
trained to estimate ambulance travel time. Conformal calibration is
subsequently used to generate prediction intervals that quantify predictive
uncertainty. Under the exchangeability assumption, the conformal methods
provide finite-sample marginal coverage guarantees. By complementing point
predictions with calibrated prediction intervals and evaluating their
empirical coverage across operationally relevant contexts, the proposed methodology
provides a basis for assessing predictive reliability beyond conventional
point prediction accuracy alone.

The key contributions of this paper are summarised as follows:
\vspace{-4pt}
\begin{itemize}
\footnotesize

\item An uncertainty-aware ambulance travel time estimation methodology that
combines domain-informed data curation, feature engineering and preprocessing,
point prediction modelling, and conformal calibration to explicitly quantify
predictive uncertainty for EMS decision support.

\item A domain-informed data curation process that establishes a more
operationally valid prediction task than the uncurated formulation and
provides a stronger basis for point prediction and subsequent uncertainty
estimation.

\item A retrospective real-world scenario study in the Skåne region of Sweden
that instantiates and evaluates the proposed methodology using two years of
ambulance trip data, including a comparison of three conformal calibration
strategies and their empirical coverage across operationally relevant
contexts.

\end{itemize}

\section{Related Work}
\label{sec:related-work}
\vspace{-9pt}
Accurate ambulance travel time estimation is fundamental to EMS dispatch
operations, as it influences ambulance allocation and compliance with
response time targets. The simplest
approaches rely on fixed-speed assumptions and lookup tables, which fail to
capture temporal and operational variability. Statistical approaches model ambulance travel times probabilistically,
for example using Bayesian data
augmentation~\cite{related-work-Bayesian-ambulance-travel-time-estimation}.
Other approaches use commercial navigation services, which systematically
misestimate lights-and-siren ambulance travel
times~\cite{introduction-ambulance-travel-time-google-maps-estimation}. More
recent machine learning methods, including gradient boosting, random forests,
and neural networks, substantially improve prediction accuracy by exploiting
GPS trajectories and spatiotemporal
features~\cite{related-work-Google-and-Data-driven-ambulance-travel-time-estimation,introduction-Adil-ambulance-travel-time,saeid-adil-simulation}.
However, these approaches almost exclusively produce deterministic point
predictions and generally assume that historical dispatch records are free
from operational artefacts.

Predictive uncertainty receives more attention in the broader travel time
prediction literature than in EMS research. Approaches that account for uncertainty include Bayesian neural networks,
quantile regression~\cite{related-work-quantile-intervals-general-travel-time-estimation},
ensemble-based prediction intervals, and, more recently, conformal
prediction~\cite{related-work-conformal-prediction-distribution-free}.
Conformal prediction provides distribution-free prediction intervals with
finite-sample marginal coverage guarantees under the exchangeability
assumption. Although it shows promise for general travel time and trajectory
prediction~\cite{related-work-conformal-intervals-general-travel-time-estimation},
its application to ambulance travel time estimation remains largely
unexplored. To address these limitations, we propose an uncertainty-aware methodology
that combines operational data curation with calibrated prediction intervals
and evaluate it through a retrospective real-world scenario study.

\section{Methodology for Uncertainty-Aware Ambulance Travel Time Estimation}
\label{sec:methodology}
\begin{figure}[t]
\centering
\includegraphics[width=\linewidth]{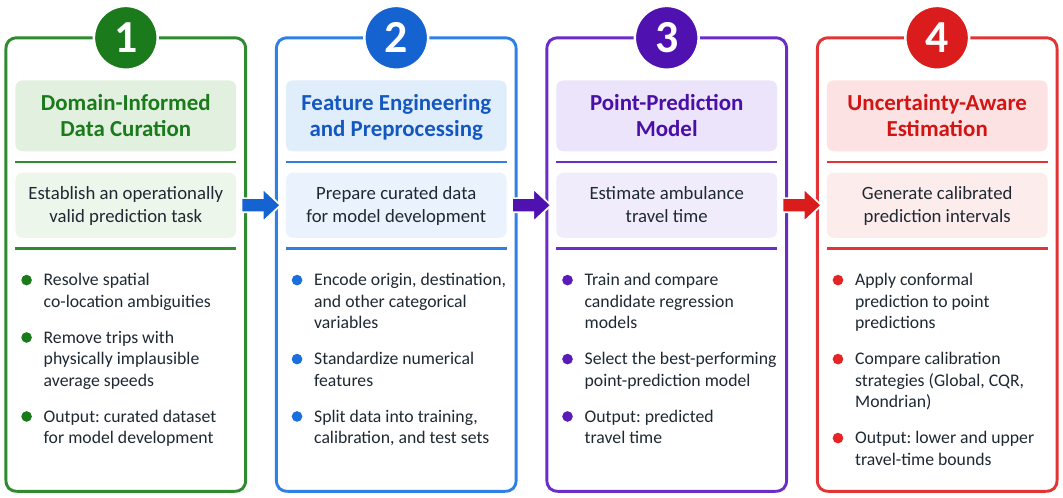}
\caption{High-level workflow of the proposed uncertainty-aware ambulance
travel time estimation methodology.}
\label{fig:method-overview}
\end{figure}

\vspace{-6pt}
Fig.~\ref{fig:method-overview} summarizes the proposed uncertainty-aware
ambulance travel time estimation methodology, which consists of four main
stages. Each stage is detailed below.

\subsection{Domain-Informed Data Curation}
\label{sec:data-curation}
\vspace{-6pt}
Real-world EMS operational data may contain recording artefacts that adversely
affect predictive accuracy and the prediction task on which subsequent
uncertainty estimation depends. Therefore, we propose performing
domain-informed data curation before model development. The first curation step addresses spatial ambiguities arising when
operationally distinct healthcare facilities share identical geographical
coordinates, such as hospitals and co-located clinics. Records affected by
these ambiguities can yield degenerate or near-zero origin--destination
separations that do not correspond to meaningful ambulance transfers. Such
records are therefore removed before model development. The second curation step addresses trips with physically implausible average speeds. For each trip, the implied average speed is computed from the
origin--destination route distance and the recorded travel time. Trips with
average speeds outside an a priori plausible range for ground ambulance
transport are removed. The plausible range should be defined based on domain
knowledge for the operational setting in which the methodology is applied.
Extremely low speeds may indicate idle or waiting time included in the
recorded trip duration, whereas implausibly low or high speeds may also result
from GPS inaccuracies or timestamp inconsistencies. For example, departure or
arrival events may be manually registered before or after the corresponding
operational event, causing the recorded trip duration to be too long or too
short. The two curation criteria are defined a priori based on EMS operational
knowledge rather than predictive performance. Together, they establish an
operationally valid prediction task for subsequent point prediction and
uncertainty estimation.

\subsection{Feature Engineering and Preprocessing}
\label{sec:preprocessing}
\vspace{-4pt}
The methodology uses information available at dispatch time to construct
features relevant to ambulance travel time estimation. These may include
spatial, temporal, operational, and route-related information. Categorical
features are represented using an encoding compatible with the selected
prediction model, while numerical features are standardized when required by
the modelling approach. All preprocessing transformations are fitted using training data only to avoid
information leakage. The resulting feature representation is then used as
input to the point prediction model. The specific implementation used in the
real-world scenario study is described in
Section~\ref{sec:experimental-setup}.

\subsection{Point Prediction Model}
\label{sec:point-prediction-model}
\vspace{-6pt}
The point prediction stage is formulated as a supervised regression task in
which ambulance travel time is estimated from dispatch-time features. The
proposed methodology is not tied to a particular regression algorithm.
Instead, one or more candidate models can be evaluated using a common
validation protocol, and the best-performing model can subsequently be used as
the point prediction component of the methodology. Model selection should be
based on predictive performance measured on data that are independent of the
final test set. The specific model candidates and model selection protocol used
in the real-world scenario study are described in
Section~\ref{sec:experimental-setup}.

\subsection{Uncertainty-Aware Ambulance Travel Time Estimation}
\label{sec:uncertainty-step}
\vspace{-6pt}

\begin{figure}[t]
\centering
\includegraphics[width=\linewidth]{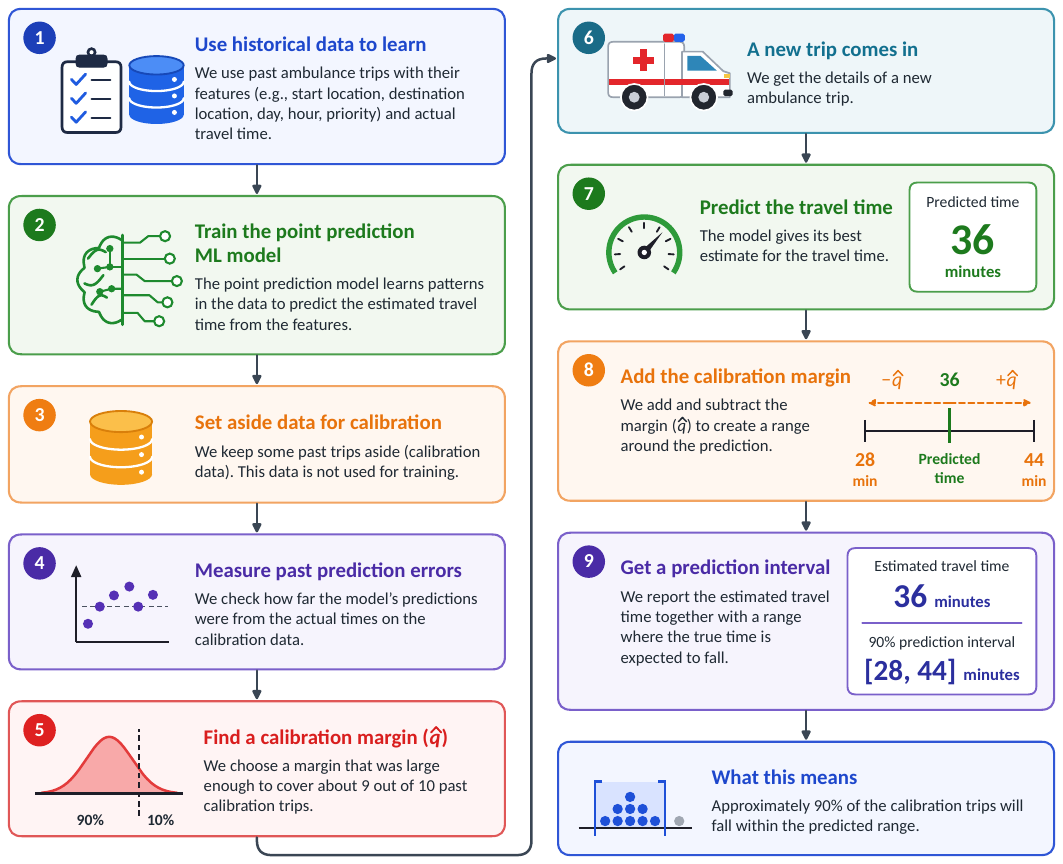}
\caption{Step-by-step workflow of global split conformal prediction for
uncertainty-aware ambulance travel time estimation.}
\label{fig:method-unertainty-overview}
\end{figure}

The point prediction model generated in stage $3$ provides a single estimated travel time for each
ambulance dispatch. However, a point estimate alone does not represent the
uncertainty associated with an individual prediction. The final stage of the
proposed methodology therefore applies conformal calibration to complement
point predictions with calibrated prediction intervals. Conformal prediction does not require a parametric assumption about the
prediction error distribution and provides finite-sample marginal coverage
guarantees under the exchangeability assumption~\cite{related-work-conformal-prediction-distribution-free}. The methodology
considers three conformal calibration strategies: global split conformal
prediction, Conformalized Quantile Regression (CQR), and priority-stratified
Mondrian conformal prediction. Fig.~\ref{fig:method-unertainty-overview}
illustrates the global split conformal workflow, comprising model development
and calibration (steps 1--5) and prediction for new dispatch requests
(steps 6--9).

\noindent{\textbf{\footnotesize Model development and calibration (steps 1--5).}}
The global split conformal workflow begins with a curated historical ambulance
trip dataset obtained through the domain-informed data curation process
described in Section~\ref{sec:data-curation}. The point prediction model $f$
is trained to estimate ambulance travel time from the dispatch-time features
(steps 1--2). To calibrate the prediction intervals, a portion of the historical trips is
reserved exclusively for calibration and is not used for model training or
selection (step~3). This \emph{calibration set},
$\mathcal{D}_{\mathrm{cal}}=\{(\mathbf{x}_i,y_i)\}_{i=1}^{n}$,
where $\mathbf{x}_i$ denotes the dispatch-time feature vector and $y_i$ the
observed travel time, is used to quantify prediction errors on observations
not used to train the point prediction model. For each calibration trip, the absolute prediction error is computed
(step~4), yielding the nonconformity score

\begin{equation}
r_i = \left| y_i - f(\mathbf{x}_i) \right|,
\qquad i = 1,\ldots,n.
\end{equation}
The final calibration step (step~5) converts these historical prediction
errors into a calibration margin, denoted by $\hat{q}$. For a target
miscoverage level $\alpha$, corresponding to nominal coverage $1-\alpha$,
$\hat{q}$ is chosen as the
$\lceil (n+1)(1-\alpha)\rceil$-th smallest value among
$r_1,\ldots,r_n$. This finite-sample-adjusted quantile, rather than the
empirical $(1-\alpha)$ quantile, provides the conformal marginal coverage
guarantee under the exchangeability assumption
~\cite{related-work-conformal-prediction-distribution-free}.

\noindent{\textbf{\footnotesize Prediction for new dispatch requests (steps 6--9).}}
For a new ambulance dispatch request (step~6), the trained point prediction
model estimates the travel time from the dispatch-time features (step~7).
The calibration margin obtained during model development is then added to and
subtracted from the point prediction (step~8), producing a calibrated
prediction interval together with the point estimate (step~9):

\begin{equation}
C(\mathbf{x})=
\left[
f(\mathbf{x})-\hat{q},
\;
f(\mathbf{x})+\hat{q}
\right].
\end{equation}

For example, a predicted travel time of $36$ minutes combined with a
calibration margin of $\hat{q}=8$ minutes produces the prediction interval
$[28,44]$ minutes (see Fig.~\ref{fig:method-unertainty-overview}).

\noindent{\textbf{\footnotesize Interpretation.}}
Under the exchangeability assumption, conformal prediction provides marginal
coverage of at least $1-\alpha$. Thus, over repeated predictions, the
proportion of future ambulance travel times contained within the reported
prediction intervals is expected to be at least the specified nominal
coverage level.

\noindent{\textbf{\footnotesize Context-aware calibration.}}
Global split conformal prediction applies the same calibration margin to every
prediction and therefore does not adapt interval width to differences in
predictive uncertainty across individual trips. To allow the methodology to
account for such differences, we additionally consider two context-aware
conformal calibration strategies. (i) Conformalized Quantile Regression (CQR)~\cite{related-work-CQR-intervals} generates trip-specific prediction
intervals from estimated lower and upper conditional quantiles. A subsequent
conformal calibration step adjusts these intervals to attain the desired
marginal coverage. (ii) Priority-stratified (Mondrian) conformal prediction~\cite{method-Mondrian conformal prediction} computes a separate calibration
quantile for each dispatch priority group and applies the corresponding margin
to new predictions within that group.

The empirical performance of global split conformal prediction, CQR, and
priority-stratified Mondrian conformal prediction is evaluated in the
real-world scenario study presented in Section~\ref{sec:results}.

\section{Real-World Scenario Study and Results}
\label{sec:results}
\vspace{-9pt}
\subsection{Scenario Description}
\vspace{-9pt}
To evaluate the proposed uncertainty-aware methodology, we instantiated it in a
retrospective real-world scenario study based on ambulance operations in the
Skåne region of southern Sweden using historical trips recorded during
2020--2021. The scenario study examines: (i)~the effect of domain-informed data
curation on point prediction performance; (ii)~the predictive performance of
alternative point prediction models; and (iii)~the calibration and
context-specific reliability of the resulting prediction intervals. The
dataset, feature configuration, candidate models, and evaluation settings
therefore represent one real-world instantiation of the proposed methodology.

\subsection{Dataset}
\label{sec:dataset}
\vspace{-4pt}
For the real-world scenario study, we used historical ambulance trip records
from the Skåne region in southern Sweden, covering the period 2020--2021. The
dataset was provided by the regional emergency dispatch authority and
contained operational information about ambulance trips. The scenario-specific input features were origin area, destination area, trip
priority, day of week, hour of day, origin--destination route distance, and a
static Google travel time estimate. The Google estimate was generated offline
and stored as a precomputed route duration attribute, requiring no live API
call during inference. Trip priority was represented using three levels: Priority~$1$ denotes
emergency trips, Priority~$2$ denotes urgent trips, and Priority~$3$ denotes
routine or less time-critical trips. The prediction target was ambulance
travel time in minutes, measured as the elapsed time between the recorded
ambulance departure and arrival timestamps. Cancelled trips and incomplete
mission segments were excluded from the dataset. For this scenario, the domain-informed data curation procedure used an
a priori plausible ground ambulance speed range of $5$--$150$ km/h.
Together with the removal of spatial ambiguities associated with co-located
healthcare facilities, the data curation procedure reduced the dataset from
$22{,}360$ original trips to $20{,}738$ operationally valid trips.

\subsection{Exploratory Data Analysis}
\label{sec:EDA}
\vspace{-6pt}
To characterize the evaluation scenario, we examined travel time variability
across operational contexts. The analysis yielded four observations that
inform the subsequent evaluation of prediction interval calibration across
operationally relevant contexts.

\begin{figure}[t]
\centering
\includegraphics[width=\linewidth]{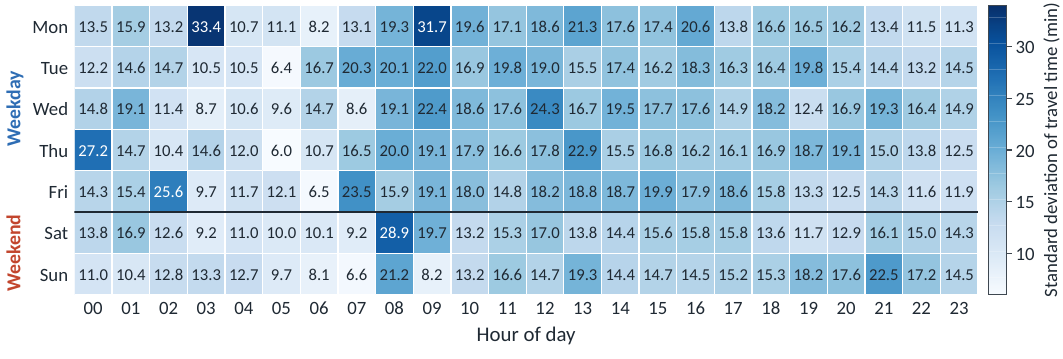}
\caption{Standard deviation of ambulance travel time by hour of day and day of
week (Darker cells indicate greater travel time variability.)}
\label{fig:eda-hour-day}
\end{figure}

\noindent{\textbf{\footnotesize Observation $1$: Travel time variability differs across operational contexts.}}
Panel~(a) of Table~\ref{tab:operational-variability} reports travel time
variability pooled across all origin--destination routes within each priority
group. These statistics therefore reflect both differences in route length
distributions and variation between trips on the same route. Relative
variability, measured using the coefficient of variation (CV), is similar
across priorities ($\mathrm{CV}=0.44$--$0.47$), although Priority~$3$ trips,
which are also the longest on average ($39.5$~min), exhibit the highest
absolute variability ($18.07$~min). Panel~(b) shows clearer differences according to route frequency. Very rare
routes have a CV of $0.592$, compared with $0.306$--$0.425$ for routes
observed more than $100$ times. Temporal variability also differs across the
day, with lower variability during nighttime hours and higher variability
during weekday mornings, particularly around $09{:}00$
(see Fig.~\ref{fig:eda-hour-day}). Variability among trips sharing similar
operational characteristics is examined further in Observation~$3$.

\begin{table}[!t]
\centering
\caption{Travel time variability across operational contexts.}
\label{tab:operational-variability}
\scriptsize

\textbf{(a) Variability by ambulance priority}

\vspace{0.3em}

\begin{tabularx}{\columnwidth}{@{}l
>{\centering\arraybackslash}X
>{\centering\arraybackslash}X
>{\centering\arraybackslash}X
>{\centering\arraybackslash}X@{}}
\toprule
Priority & Mean (min) & Std. Dev. (min) & CV & Trips \\
\midrule
Priority 1 (Emergency) & 29.79 & 13.84 & \textbf{0.465} & 2,345 \\
Priority 2 (Urgent)    & 35.24 & 15.47 & 0.439 & 3,972 \\
Priority 3 (Routine)   & 39.51 & \textbf{18.07} & 0.457 & 14,421 \\
\bottomrule
\end{tabularx}

\vspace{0.4em}

\textbf{(b) Variability by OD route frequency}

\vspace{0.3em}

\begin{tabularx}{\columnwidth}{@{}l
>{\centering\arraybackslash}X
>{\centering\arraybackslash}X
>{\centering\arraybackslash}X
>{\centering\arraybackslash}X@{}}
\toprule
OD Frequency Group & Trips & Std. Dev. (min) & IQR (min) & CV \\
\midrule
Very rare (1--5)     & 105    & 56.80 & 75.63 & \textbf{0.592} \\
Rare (6--20)         & 143    & 32.83 & 37.71 & 0.484 \\
Medium (21--100)     & 523    & 22.56 & 32.60 & 0.394 \\
Common (101--500)    & 3,871  & 15.25 & 17.89 & 0.306 \\
Very common ($>500$) & 16,096 & 14.37 & 12.92 & 0.425 \\
\bottomrule
\end{tabularx}
\end{table}

\noindent{\textbf{\footnotesize Observation $2$: Travel times exhibit a pronounced right tail.}}
The travel time distribution is strongly right-skewed. Half of all trips are
completed within $31$~min, whereas $5\%$ exceed $70$~min and $1\%$ exceed
$94$~min. Thus, long-duration trips form a non-negligible upper tail of the
observed distribution. Although the data curation procedure removes trips with
physically implausible average speeds, some long-duration observations may
still reflect delayed registration of departure or arrival events in addition
to genuine operational variability. A symmetric Gaussian approximation may
therefore inadequately represent the observed asymmetric tail of ambulance
travel times.

\noindent{\textbf{\footnotesize Observation $3$: Substantial variability remains under similar operational conditions.}}
Even when trips are grouped by the same origin--destination route and hour of
day, substantial travel time variability remains. Across $234$ groups
containing at least $20$ trips, the median within-group standard deviation is
approximately $7$~min. This remaining variability may reflect factors not
represented by the available dispatch-time features, including traffic
conditions, weather, and incident-specific circumstances. These results
indicate that trips with similar observed operational characteristics can
still exhibit substantial differences in travel time, making context-specific
evaluation of prediction intervals important in the scenario study.

\begin{table}[!t]
\centering
\caption{Empirical coverage of a naive global $95\%$ reference interval
computed from the training set mean and standard deviation.}
\label{tab:naive-coverage}

\scriptsize
\begin{tabularx}{\columnwidth}{@{}Xr@{}}
\toprule
Context & Coverage \\
\midrule
All trips (average case)      & 96.5\% \\
Morning peak (09:00)          & 94.4\% \\
Rare routes (6--20 trips)     & \textbf{69.9\%} \\
Very rare routes (1--5 trips) & \textbf{44.2\%} \\
\bottomrule
\end{tabularx}
\end{table}

\noindent{\textbf{\footnotesize Observation $4$: Aggregate coverage can conceal subgroup differences.}}
As a simple diagnostic, we applied a single global $95\%$ reference interval
across all operational contexts (Table~\ref{tab:naive-coverage}). The interval, computed as the training set mean $\pm 1.96$ standard deviations,
achieved $96.5\%$ empirical coverage overall but only $69.9\%$ for rare routes
and $44.2\%$ for very rare routes. This illustrates that satisfactory
aggregate coverage does not necessarily imply comparable coverage across
operationally relevant subgroups.

Collectively, these observations motivate evaluating prediction interval
calibration both overall and across operationally relevant contexts in the
Skåne scenario.

\subsection{Experimental Setup and Evaluation Protocol}
\label{sec:experimental-setup}
\vspace{-6pt}
The following experiments instantiate and evaluate the proposed methodology
under the Skåne scenario described above. For the point prediction stage, we
considered polynomial linear regression, random forest, CatBoost with native
categorical handling, CatBoost with one-hot encoding, and an artificial neural
network (ANN). All experiments were implemented in Python on an Apple MacBook Pro 2022
($16$-core GPU, $16$ GB RAM). 

To evaluate \emph{point prediction generalization}, we reserved $15\%$ of the
data as an independent test set. The remaining $85\%$ was used for
five-fold cross-validation, with candidate model configurations compared using
their mean validation performance across folds. After selecting the
best-performing configuration, the corresponding model was retrained on the
complete development set and evaluated once on the independent test set using
mean absolute error (MAE), root mean square error (RMSE), mean absolute
percentage error (MAPE), and the coefficient of determination ($R^2$).

For \emph{uncertainty estimation}, the same independent $15\%$ test
set was retained for final evaluation. Of the full dataset, $70\%$ was used
for training and $15\%$ for calibration. The training set was used to fit the
selected point prediction model and the quantile models, whereas the
calibration set was used exclusively to compute the conformal nonconformity
scores. Thus, the point prediction evaluation used an 85/15 development--test
split, whereas the uncertainty estimation evaluation used a 70/15/15
training--calibration--test split to preserve an independent calibration set.

The conformal calibration strategies were evaluated primarily at a nominal
coverage level of $90\%$ ($\alpha=0.10$). Calibration was additionally
examined across nominal coverage levels ranging from $50\%$ to $99\%$, and the
stability of the results was assessed across repeated random
training--calibration--test partitions.

To assess the computational feasibility of the scenario study implementation
for real-time use, we measured the end-to-end inference latency, including
feature preprocessing, point prediction, and calibrated prediction interval
generation. The implementation achieved an average single-trip inference
latency of approximately $2$ ms on the stated hardware, indicating low
computational overhead for real-time EMS decision support.

\begin{table}[t]
\centering
\caption{Effect of domain-informed data curation on point prediction performance
($\downarrow$ lower is better; $\uparrow$ higher is better).}
\label{tab:data-curation-ablation}

\scriptsize
\begin{tabular*}{\columnwidth}{@{\extracolsep{\fill}} lrrrr @{}}
\toprule
Task formulation
& \makecell{MAPE $\downarrow$\\(\%)}
& \makecell{MAE $\downarrow$\\(min)}
& \makecell{RMSE $\downarrow$\\(min)}
& $R^2$ $\uparrow$ \\
\midrule

Operationally naive (baseline)
& 12.47 & 4.83 & 10.86 & 0.62 \\

Spatially disambiguated
& 12.90 & 4.65 & 10.65 & 0.65 \\

\textbf{Operationally valid}
& \textbf{10.35} & \textbf{4.42} & \textbf{9.12} & \textbf{0.74} \\

\bottomrule
\end{tabular*}
\end{table}

\subsection{Effect of Domain-Informed Data Curation on Point Prediction Performance}
\label{sec:data-curation-ablation}
\vspace{-4pt}

We first evaluated how the domain-informed data curation stage affected
point prediction performance in the Skåne scenario. For this purpose, we
constructed three progressively curated prediction task formulations.
The \textit{operationally naive} formulation retained the data before applying
the two domain-informed curation criteria. The
\textit{spatially disambiguated} formulation additionally removed trips
affected by co-located healthcare facilities. Finally, the
\textit{operationally valid} formulation also removed trips with implied
average speeds outside the a priori plausible range of
$5$--$150$ km/h described in Section~\ref{sec:dataset}. All three formulations
were evaluated using the same ANN architecture, input features, preprocessing
procedure, and evaluation protocol. This controlled comparison isolates the
effect of progressively applying the domain-informed data curation criteria on
point prediction performance. Table~\ref{tab:data-curation-ablation} shows that the
\textit{operationally valid} formulation achieved the best overall point
prediction performance. Relative to the \textit{operationally naive}
baseline, it reduced MAPE by $17.0\%$, MAE by $8.5\%$, and RMSE by $16.0\%$,
while increasing $R^2$ from $0.62$ to $0.74$. The intermediate \textit{spatially disambiguated} formulation reduced MAE and
RMSE and increased $R^2$, although its MAPE increased slightly. This may
result from removing degenerate short trips, thereby changing the distribution
over which percentage errors are calculated. The subsequent removal of trips
with physically implausible average speeds produced the lowest prediction
errors among the three evaluated formulations. Overall, the results show that domain-informed data curation improved point
prediction performance while removing only $7.3\%$ of the original records.
This indicates that a relatively small proportion of operationally problematic
records had a substantial effect on the prediction task. These results concern
point prediction performance only and do not, by themselves, establish
improved uncertainty calibration. Prediction interval calibration is evaluated
separately in Section~\ref{sec:uncertainty-performance}.

\subsection{Point Prediction Performance}
\label{sec:point-prediction-results}
\vspace{-6pt}

Table~\ref{tab:final-model-comparison} compares the candidate point prediction
models. The ANN achieved the best overall performance, with a MAPE of
$10.35\%$, MAE of $4.42$ min, RMSE of $9.12$ min, and $R^2$ of $0.74$,
although the CatBoost models performed similarly. Polynomial linear regression
produced substantially larger prediction errors. Based on its performance
across the evaluated metrics, the ANN was selected as the point prediction
model for the subsequent uncertainty estimation experiments in this
instantiation of the proposed methodology.

\begin{table}[!t]
\centering
\caption{Comparison of candidate point prediction models
($\downarrow$ lower is better; $\uparrow$ higher is better).}
\label{tab:final-model-comparison}

\scriptsize
\begin{tabularx}{\columnwidth}{@{}l *{4}{>{\centering\arraybackslash}X}@{}}
\toprule
Model
& \makecell{MAPE $\downarrow$\\(\%)}
& \makecell{MAE $\downarrow$\\(min)}
& \makecell{RMSE $\downarrow$\\(min)}
& $R^2$ $\uparrow$ \\
\midrule

Polynomial Linear Regression
& 21.11 & 8.01 & 12.97 & 0.48 \\

Random Forest
& 13.54 & 5.36 & 10.38 & 0.67 \\

CatBoost one-hot
& 10.37 & 4.45 & 9.57 & 0.72 \\

CatBoost native
& 10.36 & 4.44 & 9.42 & 0.73 \\

\textbf{ANN}
& \textbf{10.35}
& \textbf{4.42}
& \textbf{9.12}
& \textbf{0.74} \\

\bottomrule
\end{tabularx}
\end{table}

\subsection{Evaluation of the Proposed Uncertainty-Aware Methodology}
\label{sec:uncertainty-performance}
\vspace{-4pt}
We evaluated the uncertainty estimation stage of the proposed methodology in
the Skåne scenario study by considering three evaluation questions.

\noindent{\footnotesize \textbf{Q1. Do the conformal calibration strategies achieve empirical coverage close to the desired nominal coverage?}}
We compared the three conformal calibration strategies against two
non-conformal baselines at a nominal coverage level of $90\%$
(Table~\ref{tab:conformal-overall}). The Gaussian baseline produced the widest
intervals, with a mean width of $24.7$ min, and achieved $93.2\%$ empirical
coverage.  Although its empirical coverage exceeded the nominal target, the
resulting intervals were wider than those produced by the conformal methods,
indicating lower interval efficiency. Among the conformal strategies, global split conformal prediction achieved
$89.1\%$ empirical coverage with a constant calibration margin of
$\hat{q}=8.8$ min. CQR achieved $90.5\%$ empirical coverage, while
priority-stratified Mondrian conformal prediction achieved $89.0\%$.
Thus, all three conformal strategies produced empirical coverage close to the
nominal $90\%$ target. Under the exchangeability assumption, the conformal
procedures additionally provide finite-sample marginal coverage guarantees.

\begin{table}[!t]
\centering
\caption{Comparison of uncertainty estimation methods at nominal $90\%$ coverage ($\downarrow$ lower is better).}
\label{tab:conformal-overall}

\scriptsize
\begin{tabular}{llccc}
\toprule
 & Method
 & \makecell{Coverage\\(\%)}
 & \makecell{Mean width $\downarrow$\\(min)}
 & \makecell{Median width $\downarrow$\\(min)} \\
\midrule

\multirow{2}{*}{Baselines}
 & Gaussian ($f(x) \pm 1.645\,\hat{\sigma}$)
 & 93.2 & 24.7 & 24.7 \\

 & Quantile regression (uncalibrated)
 & 89.7 & 20.4 & 19.3 \\

\midrule

\multirow{3}{*}{Conformal}
 & Global split conformal
 & 89.1 & 17.6 & 17.6 \\

 & CQR
 & \textbf{90.5} & 20.8 & 19.8 \\

 & Mondrian (priority-stratified)
 & 89.0 & 17.6 & 18.0 \\

\bottomrule
\end{tabular}
\end{table}

\noindent{\footnotesize \textbf{Q2. How well calibrated are the prediction intervals across different nominal coverage levels?}}

\begin{wrapfigure}[18]{r}{0.33\linewidth}
    \setlength{\intextsep}{1pt}
    \setlength{\columnsep}{7pt}
    \vspace{-4pt}
    \centering
    \includegraphics[width=\linewidth]{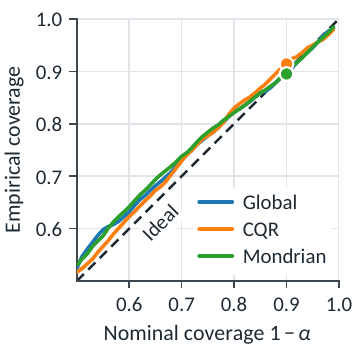}
    \caption{Calibration of empirical versus nominal coverage.
    Dashed line: perfect calibration; markers: nominal $90\%$ coverage.}
    \label{fig:calibration-curves}
    \vspace{-5pt}
\end{wrapfigure}Matching the nominal coverage at a single operating point does not necessarily
imply good calibration across other coverage levels. We therefore evaluated
empirical coverage across nominal levels ranging from $50\%$ to $99\%$
(see Fig.~\ref{fig:calibration-curves}). All three conformal strategies closely
followed the ideal calibration diagonal, indicating consistent calibration
across the evaluated range rather than only at the $90\%$ target level. This
flexibility allows the nominal coverage level to be adjusted according to the
operational requirements of the application setting. Furthermore, repeating
the evaluation across twenty random training--calibration--test partitions
yielded mean empirical coverages between $89.8\%$ and $89.9\%$ at the
$90\%$ nominal level, indicating that the deviations observed for the primary
split were small across repeated partitions.

\begin{figure}[!t]
\centering
\includegraphics[width=\linewidth]{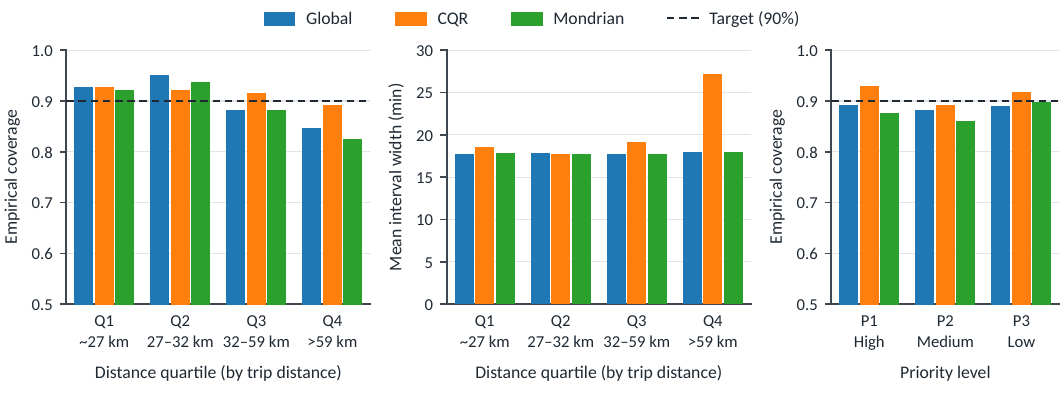}
\caption{Empirical coverage and interval width across operational contexts at nominal $90\%$ coverage.}
\label{fig:conditional-coverage}
\end{figure}

\noindent{\footnotesize \textbf{Q3. How stable is empirical coverage across operationally relevant contexts?}} The exploratory analysis in Section~\ref{sec:EDA} showed that the ambulance
travel time variability differs significantly across operational contexts. Consequently,
satisfactory marginal coverage at the aggregate level does not necessarily
imply comparable empirical coverage across different types of trips. We
therefore examined empirical coverage across dispatch priorities and trip
distance quartiles for the three conformal strategies
(see Fig.~\ref{fig:conditional-coverage}). Across dispatch priorities, all three strategies achieved empirical coverage
between approximately $86\%$ and $92\%$, indicating relatively limited
variation across the evaluated priority groups. A different pattern emerged across trip distance quartiles. Global split
conformal prediction applied a single fixed calibration margin to all trips,
whereas priority-stratified Mondrian conformal prediction applied a
priority-specific margin that remained fixed within each priority group.
Because neither strategy adapts interval width directly to trip distance, both
showed overcoverage for shorter trips and undercoverage for longer trips, with
empirical coverage falling to approximately $83\%$ for the longest-distance
quartile. This pattern is consistent with heteroscedastic travel time errors,
where predictive uncertainty increases with trip distance while the
calibration margin remains constant globally or within each priority group. In contrast, CQR generated trip-specific interval widths through conditional
quantile estimation before conformal calibration. It maintained empirical
coverage between $88.9\%$ and $91.6\%$ across the four trip distance
quartiles, while allowing interval widths to increase for trips associated
with greater predictive uncertainty.

These results show that satisfactory marginal coverage does not necessarily
imply equally stable empirical coverage across operational subgroups. Among
the three conformal strategies evaluated in the Skåne scenario study, CQR
produced the most stable empirical coverage across the trip distance quartiles
while retaining conformal marginal coverage under the exchangeability
assumption.

\section{Discussion}
\label{sec:discussion}
\vspace{-10pt}

\noindent\textbf{\footnotesize Implications for AI-Enabled Dispatch Systems.}
Traditional EMS dispatch software has relied on deterministic rules or static
lookup tables. As AI-enabled decision support methods are increasingly
introduced into EMS operations, predictive accuracy alone may be insufficient
for evaluating their reliability. By producing calibrated prediction
intervals, the proposed methodology makes predictive uncertainty explicit and
allows empirical coverage to be evaluated against a predefined target as new
operational outcomes become available. Accordingly, calibration guarantees
and empirical coverage across relevant operational contexts could complement
conventional accuracy metrics when evaluating AI-enabled dispatch support
methods for safety-critical use.

\noindent\textbf{\footnotesize Operational Self-Sufficiency.}
The implementation of the proposed methodology in the Skåne scenario study
supports self-hosted operation for two reasons. First, inference relies only
on dispatch-available features and precomputed routing attributes; no
additional data need to be queried at prediction time. Second, the point
prediction model selected for the scenario study, the ANN described in
Section~\ref{sec:experimental-setup}, is lightweight enough to run on
commodity hardware, as indicated by the inference latency measurements in
Section~\ref{sec:experimental-setup}. External routing services are used only
during offline data preparation and are not required during inference,
reducing dependence on externally hosted infrastructure. However,
incorporating real-time traffic or other contextual inputs would reintroduce
external dependencies and create a trade-off between richer predictive context
and operational self-sufficiency.

\noindent\textbf{\footnotesize Limitations and Human-in-the-Loop Governance.}
The proposed methodology is intended to support, rather than replace, human
dispatchers. Emergency dispatch remains a human-centered decision-making
process, and uncertainty-aware travel time estimates can provide additional
information that dispatchers may consider alongside contextual knowledge not
represented in the model. For example, an implementation of the methodology
cannot account for rare or unprecedented events outside the conditions
represented in the training data, such as major road closures or simultaneous
casualty incidents. Uncertainty estimates could therefore be used to identify
predictions associated with greater uncertainty and support discretionary
review rather than directly driving automated dispatch decisions. This
preserves human oversight while allowing the proposed methodology to support
routine decision-making. It also aligns with the European Union's emphasis on
human oversight of high-risk artificial intelligence
systems.

Moreover, the real-world scenario study demonstrates that the proposed methodology can be instantiated and evaluated using operational EMS data from a real regional setting. Although the evaluation was limited to the Skåne region and the 2020--2021 observation period, the methodology is designed to support adaptation to other operational contexts through context-specific data curation, feature configurations, point prediction models, and conformal calibration strategies. The present retrospective evaluation provides evidence of predictive accuracy and prediction interval calibration, including reliability across operationally relevant contexts. A natural next step is, therefore, external validation and a prospective, human-centered evaluation of how uncertainty-aware estimates affect dispatcher decisions and patient outcomes.

\section{Conclusions and Future Work}
\label{sec:conclusion}
\vspace{-10pt}

In this paper, we proposed an uncertainty-aware methodology for ambulance
travel time estimation. We evaluated the proposed methodology through a
retrospective real-world scenario study using two years of ambulance trip data
from the Skåne region in Sweden. The evaluation examined both point prediction
performance and prediction interval calibration, including empirical coverage
across operationally relevant contexts. From the evaluation results, we draw
three conclusions. First, domain-informed data curation substantially improved
point prediction performance, increasing $R^2$ from $0.62$ to $0.74$. Second,
aggregate coverage alone is insufficient to assess reliability across
operational contexts, as fixed-margin strategies undercovered the longest
trips (approximately $83\%$ coverage). Third, among the evaluated strategies,
CQR provided the most stable empirical coverage across trip distance quartiles,
maintaining $88.9$--$91.6\%$ coverage at the cost of moderately wider
intervals. These findings demonstrate the importance of evaluating both point
prediction accuracy and context-specific uncertainty when developing
trustworthy ambulance travel time estimation methods for EMS decision support.

Future work will extend the methodology by incorporating richer contextual
information, such as real-time traffic, weather, and incident-specific
characteristics.

\vspace{-2pt}

{\fontsize{9}{11}\selectfont
\paragraph*{Disclosure of Interests.}
The authors have no competing interests to declare that are relevant to the content of this article.\par}

%
%
%

\end{document}